%% file: 0-main.tex
\documentclass[letterpaper, 10 pt, conference]{ieeeconf}  
\pdfoutput=1

\IEEEoverridecommandlockouts

\usepackage{flushend}
\usepackage{balance} 

\let\labelindent\relax

\input{0-preamble}

\input{0-user-macros}

\definecolor{Lightapricot}{rgb}{0.99,0.84,0.69}

\newcommand{\modelName}{\textit{Fast}-Grasp'D\xspace}
\newcommand{\simName}{\textit{Fast}-Grasp'D\xspace}
\newcommand{\dataName}{Grasp'D-1M\xspace}

\title{\LARGE \bf
\textit{Fast}-Grasp'D: Dexterous Multi-finger Grasp Generation\\
Through Differentiable Simulation
}

\author{
Dylan Turpin$^{1,2,3}$,
Tao Zhong$^{1,2}$,
Shutong Zhang$^{1,2}$,
Guanglei Zhu$^{1,2}$,
Jingzhou Liu$^{1}$,
Ritvik Singh$^{1}$,\\
Eric Heiden$^{3}$,
Miles Macklin$^{3}$,
Stavros Tsogkas$^{1,4*}$,
Sven Dickinson$^{1,2,4}$,
Animesh Garg$^{1,2,3}$
}

\begin{document}

\twocolumn[{%
\renewcommand\twocolumn[1][]{#1}%
\maketitle
\begin{center}
\centering
\vspace{-7mm}
\includegraphics[width=0.98\linewidth]{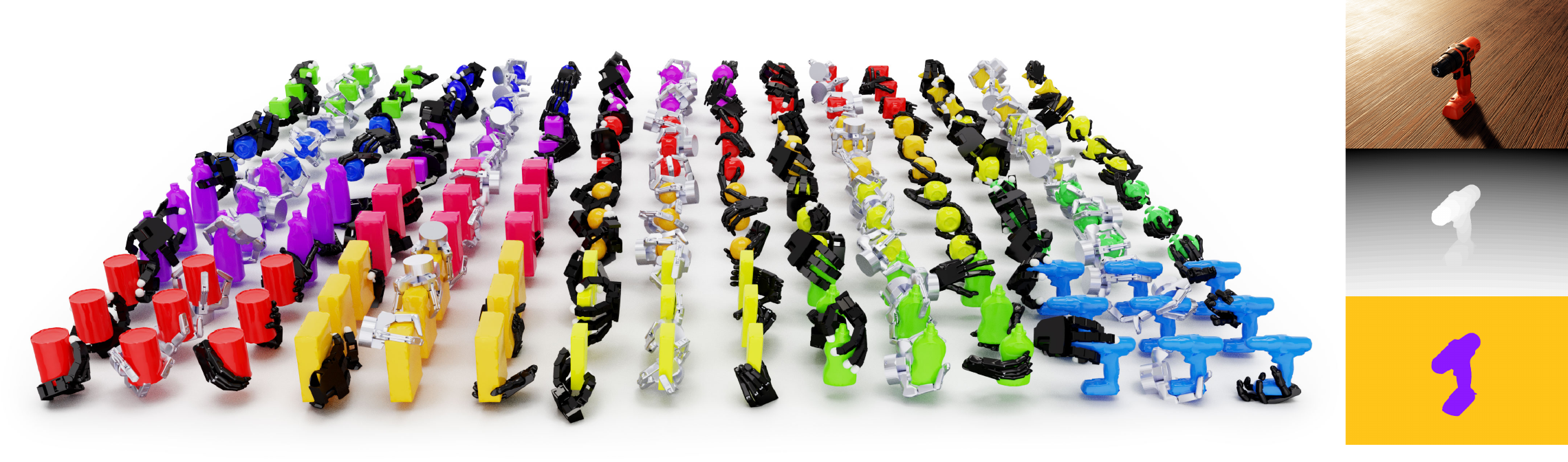}
\captionof{figure}{
\textbf{The Grasp'D-1M dataset}
contains one million unique grasps, each with multi-modal visual inputs for training vision-based robotic grasping.
We synthesize these grasps with a new differentiable grasping simulator, \simName.
Gradient information accelerates the grasp search,
allowing us to search the full-DOF space (without eigengrasps)
and simulate thousands of contacts
to produce a dataset of contact-rich, stable grasps
that can improve any learned grasping pipeline.
}
\label{fig:1-teaser}
\vspace{-5pt}
\end{center}
}]
{
  \renewcommand{\thefootnote}%
    {\fnsymbol{footnote}}
  \footnotetext[0]{
$^{1}$University of Toronto, $^{2}$ Vector Institute, $^{3}$Nvidia, $^{4}$Samsung.
\\
Correspondence: {\tt\small \{dylanturpin, garg\}@cs.toronto.edu}
\\Dickinson and Tsogkas contributed to this work in their capacity as Professor and Adjunct Professor, respectively, at the University of Toronto. The views expressed (or the conclusions reached) are their own and do not necessarily represent the views of Samsung Research America, Inc.}
}


\input{1-abstract}
\input{2-introduction}

\input{3-related-work}
\input{4-methods-v2}
\input{5-experiments}
\input{6-conclusion}

\clearpage
\renewcommand*{\bibfont}{\small}
\bibliographystyle{IEEEtran}
\bibliography{IEEEabrv,refs}

\end{document}

%% file: 0-preamble.tex
\usepackage{etoolbox} 
\usepackage[utf8]{inputenc} 
\usepackage[T1]{fontenc}    
\usepackage[english]{babel} 
\usepackage[protrusion=true,expansion=true]{microtype}
\usepackage{comment}
\usepackage{cancel}
\usepackage{csquotes}
\usepackage{listings}
\usepackage{nameref}
\usepackage{paralist}
\usepackage{xspace}
\usepackage{setspace}
\usepackage{makeidx}
\usepackage{pdfpages}
\usepackage{upgreek}

\usepackage{amsmath,amssymb} 
\usepackage{amsfonts}       
\usepackage{mathtools}
\usepackage{array} 
\usepackage{nicefrac}       
\usepackage{breqn}          
\usepackage{textcomp}
\usepackage{bm} 
\usepackage{pifont}
\usepackage[
  separate-uncertainty = true,
  multi-part-units = repeat
]{siunitx}

\usepackage{graphicx}
\usepackage{adjustbox} 
\usepackage{epsfig}

\usepackage{float}
\usepackage{subcaption}
\usepackage[font=small,labelfont=bf]{caption}
\usepackage{lscape}      

\usepackage{tikz}  
\usepackage{makecell}
\usepackage{overpic}
\usepackage{algorithm}
\usepackage[noend]{algpseudocode}

\usepackage{array} 
\usepackage{tabularx}
\usepackage{multirow}
\usepackage{multicol}
\usepackage{booktabs}   

\usepackage{paralist}
\usepackage{enumitem}
\setlist[itemize]{noitemsep,leftmargin=*,topsep=0in}
\setlist[enumerate]{noitemsep,leftmargin=*,topsep=0in}

\usepackage{url}            
\PassOptionsToPackage{table}{xcolor}
\usepackage{color,colortbl} 
\usepackage{soul}

\PassOptionsToPackage{capitalize, noabbrev}{cleveref}
\usepackage[pagebackref=false,breaklinks=true,colorlinks,urlcolor=blue,citecolor=blue,linkcolor=blue,bookmarks=false]{hyperref}

\makeatletter
\let\NAT@parse\undefined
\makeatother
\usepackage[numbers,sort&compress]{natbib}

\usepackage{blindtext}
\usepackage{bbm}

\usepackage{lipsum}

\usepackage{titlesec}
\titlespacing{\section}{0pt}{0.3\baselineskip}{0.25\baselineskip}
\titlespacing{\subsection}{0pt}{0.25\baselineskip}{0.15\baselineskip}
\titlespacing{\subsubsection}{0pt}{0.05\baselineskip}{0.03\baselineskip}

\renewcommand{\paragraph}[1]{\vspace{0.2em}\noindent\textit{#1} --}

\makeatletter
\newcolumntype{B}[3]{>{\boldmath\DC@{#1}{#2}{#3}}c<{\DC@end}}
\makeatother

%% file: 0-user-macros.tex
\renewcommand {\vec}[1]{\mathbf{#1}}

\newcommand {\x}{\vec{x}}

\newcommand {\p}{\vec{p}}
\newcommand {\pii}{\vec{p_i}}

\newcommand {\rr}{\vec{r}}

\newcommand {\rstar}{\vec{r^*}}
\newcommand {\rrhostar}{\vec{r^*_\uprho}}
\newcommand {\vi}{\vec{v_i}}
\newcommand {\vj}{\vec{v_j}}
\newcommand {\vk}{\vec{v_k}}
\newcommand {\nii}{\vec{n_i}}
\newcommand {\njj}{\vec{n_j}}
\newcommand {\nkk}{\vec{n_k}}

\newcommand\norm[1]{\Vert#1\Vert}
\newcommand {\qh}{\vec{q_h}}

\newcommand {\Lqrange}{\mathcal{L}_\textrm{range}}
\newcommand {\Lqlimit}{\mathcal{L}_\textrm{limit}}

\newcommand {\rii}{\vec{r_i}}


%% file: 1-abstract.tex
\begin{abstract}
Multi-finger grasping relies on high quality training data, which is hard to obtain:
human data is hard to transfer and synthetic data relies on simplifying assumptions that reduce grasp quality.
By making grasp simulation differentiable,
and contact dynamics amenable to gradient-based optimization, we accelerate the search for high-quality grasps
with fewer limiting assumptions.
We present \dataName: a large-scale dataset for multi-finger robotic grasping, synthesized with \simName, a novel differentiable grasping simulator.
\dataName contains one million training examples for three robotic hands (three, four and five-fingered), each with multimodal visual inputs (RGB+depth+segmentation, available in mono and stereo).
Grasp synthesis with \simName is 10x faster than GraspIt!~\cite{miller2004graspit} and 20x faster than the prior Grasp'D differentiable simulator~\cite{graspd}.
Generated grasps are more stable and contact-rich than GraspIt! grasps,
regardless of the distance threshold used for contact generation.
We validate the usefulness of our dataset by retraining an existing vision-based grasping pipeline~\cite{lundell2021ddgc} on \dataName,
and showing a dramatic increase in model performance,
predicting grasps with 30\% more contact, a 33\% higher epsilon metric, and 35\% lower simulated displacement.
Additional details at \href{http://dexgrasp.github.io}{dexgrasp.github.io}.
\end{abstract}

%% file: 2-introduction.tex
\section{Introduction}
Multi-finger robotic grasping is necessary for the effective operation of robots 
in everyday environments, which are filled with objects and affordances built for human hands.
Recent works~\cite{lundell2021ddgc,dvgg,newbury2022deep} in vision-based robotic grasping learn mappings:
(1) from visual input to gripper poses (\textit{direct regression}); or
(2) from visual input and a proposed gripper pose to a score (\textit{sampling}).
A high-quality, large-scale dataset is necessary to learn either one of these mappings.

Such data are typically generated synthetically.
Datasets of real human grasps can be captured~\cite{brahmbhatt2019contactdb,brahmbhatt2020contactpose},
and research on the grasp transfer problem
considers the best way to adapt a human grasp pose to a robot~\cite{brahmbhatt2019contactgrasp,geng2011transferring,lakshmipathy2022contact}.
However, this remains an open problem, and may be impractical in cases where
the difference between human and robot hand morphology is noticeable
(see Fig. 8 of~\cite{lakshmipathy2022contact}).
Real robot trials offer a way of evaluating proposed grasps,
but are usually considered too slow to use inside of a sampling loop.

%
%
%
%
%

Black-box optimization (e.g., simulated annealing~\cite{miller2004graspit})
takes many samples to find good grasps in the high-dimensional pose space of multi-finger grippers.
Simulation-based metrics -- widely used for parallel-jaw grasping due to their greater physical fidelity -- are too expensive to compute at each step.
Instead, we still rely largely on analytic metrics, and even then,
limit the search to a low-dimensional subspace and pre-specify a handful of possible contact locations.
\input{tab1-other-datasets}
This results in poor quality grasps.
It is unlikely that conformal grasps exist in any low-dimensional subspace we choose to search.
Simple refinement methods (e.g., \textit{autograsp} in~\cite{miller2004graspit} or the differentiable layer in~\cite{lundell2021ddgc})
create some contact by closing the fingers, but rarely discover high-contact grasps.

Our previous work~\cite{graspd} shared a similar motivation, but was impractically slow (5 minutes per grasp),
did not include data for robotic hands,
and did not evaluate whether better synthesis of training data translates to improved model performance.
%
We address these limitations with algorithmic changes to our simulator.
We use an integrator based on position-based dynamics, known to be stable and robust in contact-rich scenarios
and allowing us to forgo a problem relaxation based on contact-invariant optimization~\cite{mordatch2012discovery} (thereby reducing optimization variables from thousands to tens).
We represent the object-to-be-grasped with a mesh
(rather than a fixed sized grid),
and compute smoothed Phong~\cite{boubekeur2008phong} signed distances on the fly.
This leads to more accurate signed-distances and computation time that scales with mesh complexity.
We summarize our contributions as follows:
\begin{enumerate}
    \item We introduce the \simName simulator and pipeline for differentiable grasp synthesis ($10\times$ faster than GraspIt!~\cite{miller2004graspit} and $20\times$ faster than~\cite{graspd}) with a contact model well-suited to learning by gradient descent.
    \item We introduce the \dataName dataset of one million unique grasps with multi-modal visual input for vision-based multi-finger robotic grasping.
    \item We perform a thorough evaluation of our synthesized grasps as compared to GraspIt!, showing our grasps provide more contact and higher stability, regardless of the distance threshold used for contact generation.
    \item Finally, we demonstrate the value of \dataName by using it to improve the performance of vision-based grasp prediction, inducing 30\% more contact, a 33\% higher epsilon metric and 35\% lower simulated displacement.
\end{enumerate}

%% file: tab1-other-datasets.tex
\begin{table*}[ht!]
\begin{center}
  \begin{minipage}[c]{0.98\textwidth}
   \resizebox{\linewidth}{!}
  {
  \begin{tabular}{lcccccc}
    \toprule
    \rowcolor[HTML]{CBCEFB} 
    Dataset &
    \begin{tabular}[c]{@{}c@{}}Robotic Hands \\ (\# of Fingers) \end{tabular}&
    Visual inputs &
    Available input modalities &
    Generation method &
    \begin{tabular}[c]{@{}c@{}}Number of\\ unique grasps\end{tabular} &
    \begin{tabular}[c]{@{}c@{}}Number of\\ training examples\end{tabular} 
    \\ 
    %
    %
    %
    \midrule
    %
    \rowcolor[HTML]{EFEFEF} 
    Multi-FinGAN~\cite{lundell2020multi} & 
    Barrett (3) &
    \checkmark &
    RGBD, segmentation &
    GraspIt!~\cite{miller2004graspit} &
    1,355 &
    4,990\\
    %
    DDGC~\cite{lundell2021ddgc}&
    Barrett (3) &
    \checkmark &
    RGBD, segmentation &
    GraspIt!~\cite{miller2004graspit} &
    6,793 &
    185,598\\
    %
    \rowcolor[HTML]{EFEFEF} 
    Columbia Grasp Database~\cite{goldfeder2009columbia}& 
    Barrett (3) &
    \text{x}&
    (none) &
    GraspIt!~\cite{miller2004graspit} &
    158,006 &
    (none)\\
    \midrule
    \textbf{\dataName} (ours)&
    \begin{tabular}[c]{@{}l@{}}\textbf{Barrett (3), Allegro (4)} \\ \textbf{Shadow (5)} \end{tabular}&
    \checkmark &
    \begin{tabular}[c]{@{}c@{}}\textbf{RGBD, segmentation,} \\ \textbf{2D/3D bbox (in mono+stereo)} \end{tabular}&
    \begin{tabular}[c]{@{}c@{}}Differentiable \\ Simulation \end{tabular}&
    \textbf{1,000,000} &
    \textbf{1,000,000}\\
    \bottomrule
  \end{tabular}
  }
  \end{minipage}
\end{center}
\caption{\textbf{Datasets of multi-finger robotic grasps} for training
vision-based grasping are uncommon, limited in size (especially when considering the number of unique grasps, which are reused with multiple scenes or camera angles)  and
contain grasps whose quality is limited by the assumptions necessary for sampling-based planning with GraspIt!~\cite{miller2004graspit} to succeed.}
\label{tab:1-other-datasets}
\end{table*}

%% file: 3-related-work.tex
\section{Related Work}

\paragraph{\textbf{Vision-based grasp prediction}}
Modern approaches~\cite{newbury2022deep,zhang2022robotic} to grasp prediction learn a mapping from visual inputs
to grasps
(or to a quality function used alongside a sampler)
by training on a dataset of positive examples.
~\cite{corona2020ganhand,lundell2020multi,lundell2021ddgc} employ GAN-style models to predict stable grasps from RGBD inputs.
~\cite{karunratanakul2020grasping,khargonkar2022neuralgrasps} take an implicit approach to
grasp representation
by learning to jointly predict signed distances for a gripper and
object to be grasped.
Recent works on parallel-jaw grasping~\cite{mahler2019learning,mousavian20196,sundermeyer2021contact, jiang2021synergies}
use datasets derived from simulation~\cite{kappler2015leveraging,eppner2021acronym},
which have better physical fidelity~\cite{danielczuk2019reach,mousavian20196,eppner2021acronym}
and more intuitive plausibility~\cite{kappler2015leveraging} than analytic datasets.
In contrast, multi-finger robotic grasp prediction continues to be trained on analytically synthesized datasets.
\cite{lundell2020multi,lundell2021ddgc,karunratanakul2020grasping,khargonkar2022neuralgrasps,hasson2019learning,doosti2020hope,jiang2021hand} all use analytically synthesized datasets from the GraspIt! simulator~\cite{miller2004graspit}.
We aim to improve vision-based multi-finger grasping
by using simulation to generate better datasets
(and using gradient-based optimization to make the higher computational cost of simulation affordable).

\paragraph{\textbf{Multi-finger robotic grasping datasets}}
Only a handful of datasets exist for multi-finger grippers (see Table~\ref{tab:1-other-datasets}).
Of these, most include only the three-finger Barrett hand~\cite{goldfeder2009columbia,lundell2020multi,lundell2021ddgc}
or do not include visual inputs to learn from~\cite{goldfeder2009columbia,brahmbhatt2019contactgrasp,shao2020unigrasp}.
The \dataName dataset contains grasps for grippers with three (Barrett), four (Allegro), and five (Shadow) fingers,
along with a variety of multi-modal visual inputs to learn from.
Furthermore, through differentiable simulation, we are able to synthesize
more physically-plausible, stable, and contact-rich grasps
than can be found with the analytic grasp synthesis~\cite{miller2004graspit,pokorny2013classical} used to generate other datasets.

\noindent \textbf{Grasp synthesis.}
Since human grasps are difficult to transfer to robotic hands,
and gathering real robot data is expensive (and presupposes a way to generate trial grasps),
robotic grasping datasets often rely on artificial grasp synthesis.
Analytic synthesis, which optimizes a handcrafted metric to find stable grasps,
has been successfully applied to \textit{parallel-jaw} grippers
(based on grasp wrench space
analysis~\cite{ferrari1992planning,miller2004graspit,goldfeder2009columbia},
robust grasp wrench space analysis~\cite{weisz2012pose,mahler2017dex},
or caging~\cite{rodriguez2012caging,mahler2016energy}).
%
While they are more computationally costly, simulation-based metrics better align with human judgement~\cite{kappler2015leveraging, depierre2018jacquard} and with real world
performance~\cite{mahler2019learning,danielczuk2019reach,mousavian20196,eppner2021acronym}.
Unfortunately, this higher computational cost has delayed the adoption of simulation-based
synthesis for multi-finger hands.
Sample-intensive black-box optimization in a high-dimensional pose space
renders simulated metrics too expensive.
In fact, for high-DOF hands, optimizing over even simple analytic metrics is
usually impractical without limiting search to a low-dimensional subspace~\cite{ciocarlie2007dexterous}
and considering only a handful of pre-specified contact locations.
%
%
%
%
We introduce a differentiable simulation-based metric.
Gradient-based optimization reduces the number of samples required,
and GPU parallelism makes simulation fast enough
that we can search the full grasp space and simulate thousands of contacts
in order to find contact-rich, physically plausible grasps.


\begin{figure*}
    \centering
    \begin{minipage}{0.7\linewidth}
    \includegraphics[width=\linewidth]{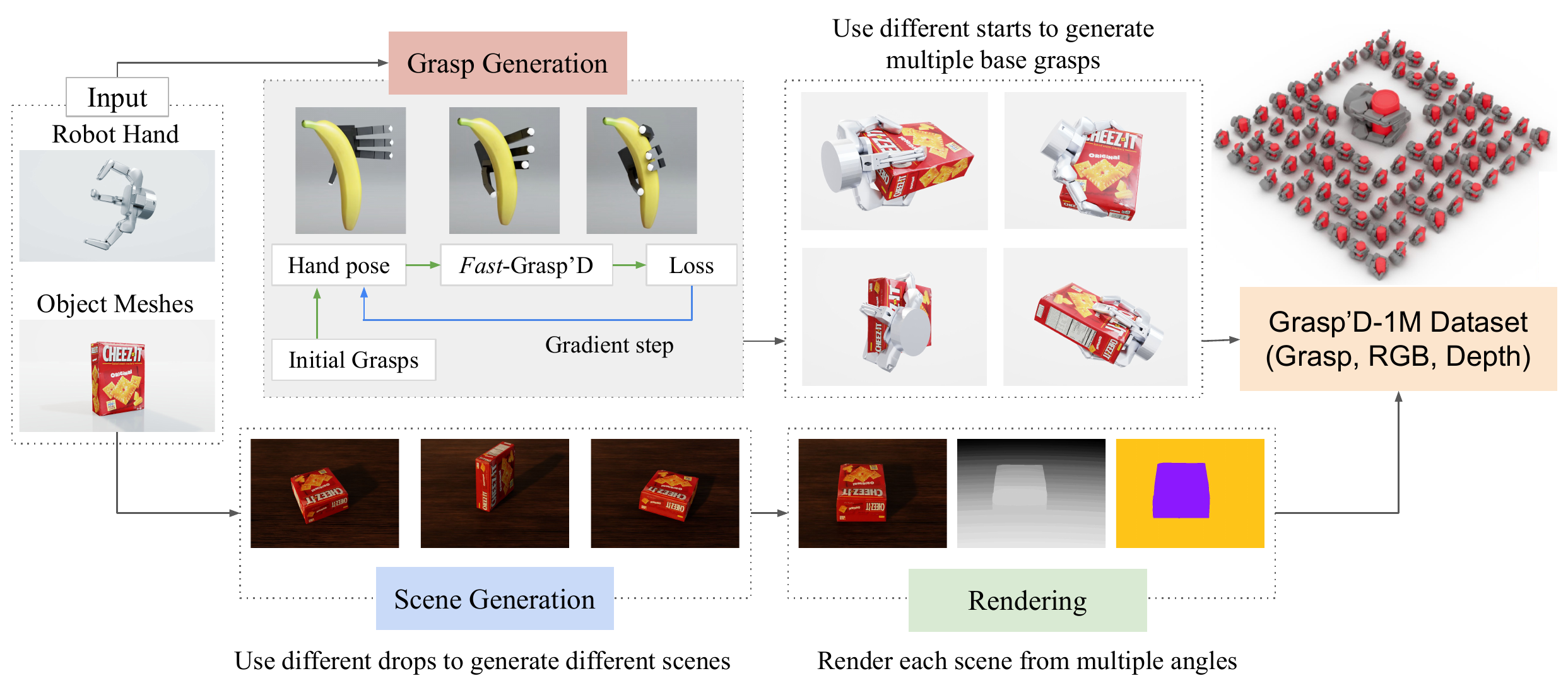}
    \end{minipage}
    \begin{minipage}{0.29\linewidth}
    \caption{\textbf{Our grasp synthesis pipeline} generates the \dataName dataset of one million unique grasps in three stages.
    \textit{(1) Grasp generation}: For any provided $(\textrm{robot hand}, \textrm{object})$ pair, we generate a set of base grasps by gradient descent over an objective computed by \simName, our fast and differentiable grasping simulator.
    \textit{(2) Scene generation}: We simulate multiple drops of each object onto a table to create scenes with different object poses and transfer base grasps to these scenes.
    \textit{(3) Rendering}: Finally, we render each scene (RGB, depth, segmentation, 2D/3D bounding boxes in mono+stereo) from multiple camera angles.
    }
    \label{fig:2-method}
    \end{minipage}
\end{figure*}

\paragraph{\textbf{Differentiable Physics}}
While there has been brisk recent progress in differentiable 
physics engines~\cite{hu2019chainqueen, hu2019difftaichi, geilinger2020add, brax2021github, werling2021fast, qiao2021efficient, heiden2021neuralsim, heiden2021disect, xie2022shac}, myriad limitations 
render them inadequate for our application.
Brax~\cite{brax2021github} and the Tiny Differentiable Simulator~\cite{heiden2021neuralsim} only support collision primitives and cannot model collisions between complex meshes.
Nimblephysics~\cite{werling2021fast} supports mesh-to-mesh collision, but cannot handle cases where the gradient of
contact normals with respect to position is zero (e.g., on a mesh face).
While its analytic computation of gradients is fast, Nimblephysics requires the manual derivation of the backward pass when new simulation functionality is added. Instead, in this work, similar to Brax~\cite{brax2021github} and Taichi~\cite{hu2019difftaichi}, we rely on automatic differentiation to translate user-defined simulation code from high-level Python to kernel codes that run very efficiently on GPUs.

\paragraph{\textbf{Differentiable Grasping}}
Differentiable multi-finger grasp synthesis is a less explored domain.
\cite{liu2020deep} and \cite{liu2021synthesizing} formulate differentiable
force closure metrics and use gradient-based optimization to synthesize grasps with
the Shadow and MANO~\cite{MANO:SIGGRAPHASIA:2017} hands, respectively.
This allows analytic synthesis to search the full-dimensional pose space of a high-DOF hand, yet still exhibits the usual drawbacks of analytic metrics.
\cite{liu2020deep} assumes that contact is limited to 45 manually labelled points,
and \cite{liu2021synthesizing} assumes zero friction,
uniform force magnitude across all contacts,
and only scales to grasps with a few point contacts
(with three contacts it takes $\sim40$ minutes to find 5 acceptable grasps).
Our previous work~\cite{graspd}
formulated a differentiable simulation-based metric
able to scale to thousands of contacts to approximate surface contact
and generate plausible, contact-rich grasps.
However, Grasp'D~\cite{graspd} was still slow
($\sim5$ minutes per grasp),
focused on human rather than robotic hands,
did not release a dataset or quantitative evaluation for robot grippers,
and did not evaluate the benefit of actually training vision-based pipelines on the generated data.
\simName addresses these concerns with
up to a $30\times$ speedup ($\sim$10s per grasp online or $\sim$1s amortized),
a large-scale dataset for vision-based robotic grasping,
and a thorough evaluation of an existing vision-based grasping pipeline retrained
using \modelName data.
Two concurrent works~\cite{li2022gendexgrasp,wang2022dexgraspnet} also propose differentiable stability metrics, but their fidelity is limited by considering only a handful of contact locations at a time.
We include a favourable comparison to~\cite{li2022gendexgrasp} in Section~\ref{sec:experiments}.
\cite{wang2022dexgraspnet} has not yet released data.

%% file: 4-methods-v2.tex
\section{\simName: Differentiable Grasp Synthesis}
\label{sec:method-grasp-synthesis}
We present a method for grasp synthesis 
using an input object mesh and hand model (represented by a mesh and an articulation chain with meshes at the links),
and generate a physically-plausible stable grasp as a base pose and joint angles of the hand.
This is achieved by gradient descent over a stability metric computed by our differentiable grasping simulator, \simName.
The final grasp is dependent on the pose initialization of the hand,
so different grasps can be recovered by sampling different starting poses.
We extend differentiable grasp synthesis from previous work~\cite{graspd} 
with algorithmic changes to achieve a $20\times$ speedup in performance, generating a grasp in about one second.
Namely, we improve the method's integration scheme
(to ensure stable optimization without introducing additional relaxation variables)
and 
object representation
(to scale computation with object complexity).
We also adapt the concepts of signed-distance function dilation and \textit{leaky gradient} to position-based dynamics.
We build on the Warp~\cite{warp} framework, which supports fast auto-differentiation and GPU acceleration.

\subsection{From Grasp'D to \simName}
We outline the challenges that motivate our design.
\paragraph{\textbf{Gradient-friendly contact dynamics for mesh inputs}}
The algorithms usually employed for mesh-to-mesh collision rely on operations
(e.g., tree-traversal) that are hard to differentiate through.
A formulation of contact constraints based on signed distance functions (SDFs) is well-suited to differentiable collision detection~\cite{macklin2020local,graspd}.
This leads us to two requirements:
i) we need a way to compute and represent an SDF based on a mesh;
ii) since the true SDF surface is locally flat, we need a way of smoothing it.
Grasp'D~\cite{graspd},
addressed these requirements by pre-computing a discretized SDF grid
from which values were computed by trilinear interpolation, which acts as a simple form of smoothing.
Storing and querying the grid has a constant memory and compute cost.
Surface normals can be computed (differentiably) by finite differencing.
On the other hand, the grid is an approximation that loses details from the underlying mesh
and under constant voxel resolutions is not cheaper to use with simpler meshes.

In this work, we compute the object SDF directly on the triangular, watertight mesh representing the object. We leverage the bounding volume hierarchy (BVH) data structure to accelerate the query of the closest mesh face to compute the distance, and use ray casting to determine the sign of the result (negative sign means inside, positive sign means outside the mesh).
Since we compute the true SDF and not an approximation,
%
local flatness of the mesh -- i.e., constant normals along each face --
creates zero-gradient plateaus that are hard to escape.
To address this, we take inspiration from Phong tessellation~\cite{boubekeur2008phong}, a rendering technique that
smooths mesh normals and silhouettes,
and computes a smoothed \textit{Phong SDF}.
%
This is more accurate than discretizing, 
requires $7\times$ fewer SDF queries (by avoiding finite differencing for normals),
and lets computation time scale with mesh complexity, leading to a large speedup when using simplified meshes
provided by DDGC~\cite{lundell2021ddgc}.
Discontinuities between faces --
where normals vary sharply along an edge --
create unstable regions where gradient steps produce unexpected
results.
Here we follow~\cite{graspd} and take a coarse-to-fine smoothing approach (see Section~\ref{sec:method-sdf-dilation}).

\paragraph{\textbf{Instability in integration and optimization}}
Our grasp synthesis method consists of an inner loop (integration)
that computes a simulation-based grasp metric maximized
by an outer loop (optimization).
Study of the outer-loop optimization properties of different differentiable integration schemes is just beginning~\cite{suh2022differentiable,zhong2022differentiable}.
Inner-loop instability may arise from simulating hard contact constraints with a force-based integrator.
%
%
Enforcing non-penetration between rigid bodies under a force-based integrator requires high-stiffness constraints.
Such constraints may cause instability, necessitating short time steps,
since small pose changes (inducing small constraint violating interpenetrations) create large forces.
Instability in the inner loop (integration) destabilizes the outer loop (optimization),
creating a rugged loss landscape that is hard to optimize over, even with gradient information.

Our previous work~\cite{graspd} handled this instability with a problem relaxation that allowed (minimal) physics violations.
Inspired by Contact-Invariant Optimization~\cite{ciocarlie2007dexterous,ciocarlie2009hand}, 
each contact point was assigned a corresponding six-dimensional variable representing the desired resultant object wrench.
%
%
Instead, our current approach (see Section~\ref{sec:method-pbd})
is built on position-based dynamics (PBD~\cite{muller2007position}), known for stability and robustness in contact-rich scenarios.
%
%
A more stable inner loop allows grasp synthesis to succeed without introducing additional variables, leading to a significant speed increase (about $20\times$ over~\cite{graspd}).

\paragraph{\textbf{Contact sparsity}}
Most points on the hand are not in contact with the object
and will not be brought into contact by an infinitesimal pose perturbation.
This means most hand vertices do not contribute to the simulator gradient, so it is difficult for gradient-descent to create new contacts.
To address this, we adapt the \textit{leaky} contact gradients of~\cite{graspd} to PBD.

\subsection{Shape representation}
\paragraph{\textbf{Signed distance function (SDF)}}
Whereas primitive objects (e.g., a sphere or box) admit an analytic SDF,
this is not the case for complex objects, for which an SDF representation is not readily available.
We represent the object to be grasped by a mesh in canonical pose, from which we compute SDF values
on the fly as $\phi(\rr) = \rr^{\pm}_{\textrm{obj}} \norm{\rr - \rstar}$,
where $\rstar$ (the closest point on the mesh surface to $\rr$)
and $\rr^{\pm}_{\textrm{obj}}$ (a positive/negative sign indicating whether $\rr$ is outside/inside the mesh volume)
are queried from Warp.

\paragraph{\textbf{Phong SDF}}
\label{sec:method-phong}
As described in the introduction of this section,
we use the smoothed \textit{Phong SDF} $\rho(\rr)$ for contact generation~\cite{boubekeur2008phong}.
The \textit{Phong SDF} is the SDF of a quadratic surface matching
normals interpolated from the vertices of the face $\rstar$ lies on according to barycentric coordinates.
Say $\rstar = u\vi + v\vj + w\vk$, with $u,v,w$ being the barycentric coordinates
and $\vi,\vj,\vk$ being the vertices
with vertex normals $\nii,\njj,\nkk$.
We then define the closest point to $\rr$ on the quadratic surface as
\begin{align*}
\rrhostar = (1 - \alpha) \alpha (u, v, w)\Big(\begin{smallmatrix}
  \psi_i(\rstar)\\
  \psi_j(\rstar)\\
  \psi_k(\rstar)
\end{smallmatrix}\Big),
\end{align*}
where $\psi_i(\rstar)$ is the projection of $\rstar$ onto
the plane defined by $\vi$ (and $\nii$),
and $\alpha$ controls interpolation between the flat
and curved triangle.
Finally, we may write our \textit{Phong SDF} as $\rho(\rr) = \rr^{\pm}_{\textrm{obj}} \norm{\rr - \rrhostar}$, with
${\nabla \rho}(\rr) = (\rr - \rrhostar) / \norm{\rho(\rr)}$.

\paragraph{\textbf{SDF dilation}}
\label{sec:method-sdf-dilation}
In addition to \textit{Phong SDF} smoothing,
we follow the coarse-to-fine smoothing introduced in~\cite{graspd}.
Specifically, we define the object surface $\emph{not}$ as 
the zero-level of the SDF $\rho$, but as the radius $r \geq 0$ level-set,
which yields a padded, rounded version of the surface.
We still want the final grasp to respect the real object geometry,
so we decrease $r$ to $0$ on a linear schedule, gradually resolving
the dilated SDF to the true surface as optimization continues.

\subsection{Position-based dynamics}
\label{sec:method-pbd}
We represent the hand by its vertices, $\p = \textrm{FK}(\qh)$, whose positions are given by applying differentiable forward kinematics $\textrm{FK}$ to the hand configuration $\qh$.
Each hand vertex $\pii$ imposes a non-penetration constraint
from which we compute positional updates to the object
orientation (we treat the hand as static).
Let $\x$, $\theta$ and $\dot{\x}$, $\dot{\theta}$,
be the object position, orientation, and their time derivatives.
We can express each hand point in terms of its location in the object local frame as
$\rii = R(\theta)^{-1}(\pii - \x)$, where $R(\theta)$ is the object rotation matrix.
The non-penetration constraint can then be written as
$C(\pii) = \max(0, -\rho(\rii))$,
which computes penetration depth using the object \textit{Phong SDF}.
Given current values for $\x^{-},\dot{\x}^{-},\theta^{-},\dot{\theta}^{-}$,
to perform an integration step,
we first compute predicted values with a symplectic Euler update (see preliminaries in~\cite{deul2016position}), yielding
$\tilde{\x},\tilde{\dot{\x}},\tilde{\theta},\tilde{\dot{\theta}}$.
Next, we compute updates based on each constraint as:
\begin{equation}
    \label{eq:update}
    \left[\nabla \x^T \nabla\theta \right]^T = -\vec{M}^{-1}\vec{J}_C^T \left[\vec{J}_C \vec{M}^{-1} \vec{J}_C^T \right]^{-1}C(\pii),
\end{equation}
where $\vec{M}$ is the mass matrix and $\vec{J}_C(\x,\theta) = \left( \frac{\partial C}{\partial \x} \frac{\partial C}{\partial \theta}\right)$ is the constraint Jacobian.
We perform a Gauss-Seidel step to update the predicted values for all constraints, yielding $\x^+,\theta^+$, and
set derivatives accordingly as $\dot{\x}^+ = (\x^+ -\x^-)/\Delta t$ and $\dot{\theta}^+ = (\theta^+ -\theta^-)/\Delta t$, where $\Delta t$ is the timestep length.
We follow the Coulomb friction formulation from~\cite{deul2016position}.

\begin{figure}[!t]
    \centering
    \includegraphics[width=\linewidth]{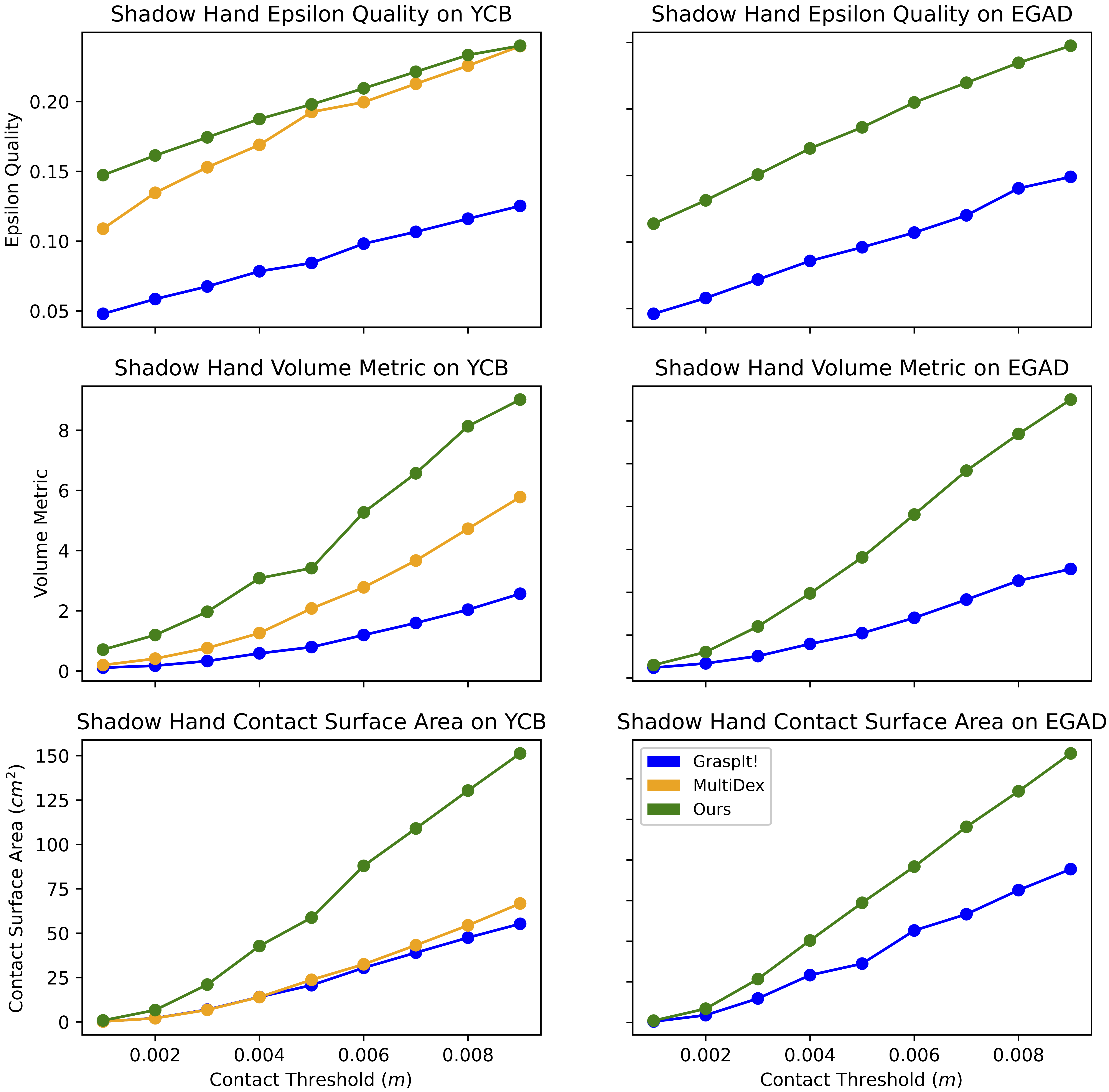}
    \caption{\textbf{Grasp metrics} such as epsilon quality, GWS volume, and contact surface area depend on the threshold distance used for contact generation.
    Our method improves on GraspIt!~\cite{miller2004graspit} and MultiDex~\cite{li2022gendexgrasp} baselines under all threshold choices.
    Results for the Barrett and Allegro hands (available on our website) follow a similar trend.
    }
    \label{fig:3-grasp-metrics}
\end{figure}

\paragraph{\textbf{Leaky gradient}}
\label{subsec:contact-dynamics}
The non-penetration constraint shows that a hand vertex (say $\pii$ and $\rii$ in world and object frame) not in collision with the object (i.e., with $\rho(\rii) > 0$)
will have $C(\pii) = 0$ and,
from equation \eqref{eq:update}, will not contribute to the PBD update.
A small perturbation of hand pose will not create contact ($\rho(\rii + \epsilon) > 0$), so the vertex will not contribute to hand pose gradient.
This makes it difficult to follow gradient to create new contacts.
We address this with the \emph{leaky gradient} introduced in~\cite{graspd}.
Specifically, rather than using the correct gradient $\frac{\partial C}{\partial \p} = \begin{cases}
\nabla \phi(\p) &\textrm{if $\phi(p) < 0$}\\0 &\textrm{otherwise}
\end{cases}$,
we use a biased gradient
$\frac{\partial C}{\partial \p} = \begin{cases}
\nabla \phi(\p) &\textrm{if $\phi(\p) < 0$}\\\alpha\nabla \phi(\p) &\textrm{otherwise}
\end{cases}$,
where $\alpha \in [0,1]$ controls how much gradient leaks through.
We set $\alpha=0.1$ in our experiments. 

\section{\dataName: Dexterous Grasp Dataset}
\label{sec:method}

\simName enables an algorithmic pipeline for generating multi-finger grasps.
Given object and gripper sets,
we synthesize many grasps for each $(\textit{gripper},\textit{object})$ pair
with the object set in canonical pose.
We call these \textit{base grasps}.
Next, we generate several scenes for each object
by dropping the object (in simulation) on a table and letting it come to rest in a natural pose.
We transfer \textit{base grasps} to scenes
by applying the object pose transform to the gripper pose,
yielding a larger set of \textit{scene grasps}.
Finally, we render each \textit{scene} from multiple viewpoints,
yielding a set of training examples that can be used to 
train vision-based grasping.

\subsection{Gradient-based Grasp Optimization}
\label{sec:grasp-metric}

\paragraph{\textbf{Computing the grasp metric}}
To measure the quality of a candidate grasp $\qh$,
we test its ability to withstand forces applied to the object.
Specifically, we set an initial object velocity  ${\dot{\x}}^{(0)}$ and test whether contact with the static gripper can dampen it.
The object is always initialized in canonical pose $(\x^{(0)}, \theta^{(0)})$.
We simulate according to Section~\ref{sec:method-pbd}
and compute the object's final (translational and angular)
velocity $({\dot{\x}}^{(T)}, \dot{\theta}^{(T)})$.
The more the velocity is dampened, the more stable we estimate the grasp to be.
Of course, testing a single force is not sufficient
and we perform a batch of $M$ simulations in parallel,
each setting a different initial object velocity $\dot{\x}^{(0)}_m$.
In our experiments we use $M=7$, setting positive and negative velocities along each axis as well as one simulation with $\dot{\x}^{(0)} = 0$. We simulate for a single timestep, with $\Delta t = 0.001s$.
Our stability loss is then defined as 
\begin{equation}
    \mathcal{L}_{\textrm{stable}}(\qh) = \sum_{m=1}^M \frac{\norm{\dot{\x}^{(T)}_m} + \norm{\dot{\theta}^{(T)}_m}}{M}.
\end{equation}

\paragraph{\textbf{Additional losses}}
$\Lqrange$ encourages hand joints to be near the middle of their ranges.
$\Lqlimit$ penalizes hand joints outside of their range.
\begin{align}
    \Lqrange(\qh) &= \norm{\qh - \frac{ \qh^{\textrm{up}} + \qh^{\textrm{low}}}{2}}\\
    \Lqlimit(\qh) &= \max(\qh - \qh^{\textrm{up}}, 0) + \max(\qh^{\textrm{low}} - \qh, 0)
\end{align}

\paragraph{\textbf{Optimization}}
We sample hand initializations $\qh$ with the approach-sampling  procedure described in~\cite{graspd}.
We optimize using simple gradient descent with a learning rate of $0.001$.

\begin{figure}[!t]
    \centering
    \includegraphics[width=\linewidth]{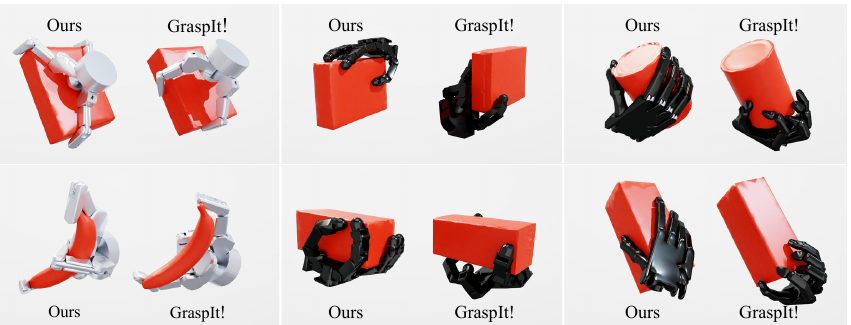}
    \caption{\textbf{Contact-rich grasps} can be generated by our method which optimizes in the full DOF-space of the hand. The GraspIt!~\cite{miller2004graspit} planner mainly generates fingertip grasps. The baseline grasps exhibit fewer contacts that result in reduced stability compared to grasps synthesized by our method.}
    \label{fig:qualitative}
\end{figure}

\input{5-expt-tables}

\paragraph{\textbf{Dataset generation}}
We use the Omniverse Replicator to generate and render scenes as visual input for learning.
Each scene has an object on a table.
To generate the scene, we sample an object pose above the table and simulate dropping the object and letting it come to rest. 
%
The renders include RGB, depth, 2D and 3D object bounding boxes and a segmentation (separating the image into labelled regions for
the table, object and background) in both mono and stereo.
We generate training tuples by transferring base grasps to each scene.
Specifically we apply the scene's object pose transform to the base grasp and check for interpenetration with the table.

%% file: 5-expt-tables.tex
\begin{table*}[!t]
  \centering
  \begin{minipage}{0.6\linewidth}
  \resizebox{\linewidth}{!} 
  {
  \begin{tabular}{lll|S[table-format=4.5]S[table-format=4.5]S[table-format=4.5]S[table-format=4.5]S[table-format=4.5]}
  \toprule
  \rowcolor[HTML]{CBCEFB} 
  \textbf{Objects} &\textbf{Hand} &\textbf{Dataset} &  \textbf{CA} $\uparrow$ & \textbf{IV} $\downarrow$ & \textbf{{\Large$\epsilon$} $\uparrow$} & \textbf{Vol $\uparrow$} & \textbf{SD $\downarrow$} \\ 
  \midrule
  \rowcolor{Lightapricot}
  Scale (Unit) & & & \small{$\text{cm}^2$} & \small{$\text{cm}^3$} &  &  & \small{$\text{cm}$} \\ 
  \midrule
  YCB & Barrett & \emph{DDGC}~\cite{lundell2021ddgc} & 4.64 & 2.73 & 0.10 & 0.26 &2.74\\
  \rowcolor[HTML]{EFEFEF} 
  & & \emph{GraspIt!}~\cite{miller2004graspit} & 7.18 & \textbf{0.54} & 0.13 & 0.46 & 2.10 \\
  & & \emph{MultiDex}~\cite{li2022gendexgrasp} & 13.82 & 5.55 & 0.18 & 1.38 & \textbf{1.24} \\
  \rowcolor[HTML]{EFEFEF} 
  && \emph{Ours} & \textbf{55.71} & 4.70 & \textbf{0.23} & \textbf{5.09} & 1.40 \\
  & Allegro & \emph{GraspIt!} & 12.37 & \textbf{1.80} & 0.15 & 1.80 & 2.01 \\
  \rowcolor[HTML]{EFEFEF} 
  & & \emph{MultiDex} & 21.16 & 6.61 & \textbf{0.18} & 2.19 & 1.93 \\
  & & \emph{Ours} & \textbf{49.17} & 5.62 & \textbf{0.18} & \textbf{3.42} & \textbf{1.66} \\
  \rowcolor[HTML]{EFEFEF} 
  & Shadow & \emph{GraspIt!} & 20.72 & 6.68 & 0.08 & 0.80 & 2.88 \\
  & & \emph{MultiDex} & 21.47 & 6.70 & 0.17 & 1.90 & 2.60 \\
  \rowcolor[HTML]{EFEFEF} 
  & & \emph{Ours} & \textbf{58.60} & \textbf{6.34} & \textbf{0.18} & \textbf{3.42} & \textbf{2.42} \\
  \midrule
  EGAD & Barrett  & \emph{DDGC} & 0.69 & \textbf{0.02} & 0.05 & 0.07 & 2.84 \\
  \rowcolor[HTML]{EFEFEF}
  & & \emph{GraspIt!} & 10.96 & 2.13 & 0.14 & 0.81 & 1.68 \\
  & & \emph{Ours} & \textbf{58.17} & 4.80 & \textbf{0.24} & \textbf{3.71} & \textbf{0.99} \\
  \rowcolor[HTML]{EFEFEF} 
  & Allegro & \emph{GraspIt!} & 14.38 & \textbf{1.98} & 0.16 & 1.98 & 1.57 \\
  & & \emph{Ours} & \textbf{36.68} & 6.54 & \textbf{0.22} & \textbf{4.49} & \textbf{1.05}\\
  \rowcolor[HTML]{EFEFEF} 
  & Shadow & \emph{GraspIt!} & 28.96 & 8.91 & 0.12 & 1.61 & 3.27 \\
  & & \emph{Ours} & \textbf{58.91} & \textbf{5.18} & \textbf{0.24} & \textbf{7.05} & \textbf{1.71} \\
  \bottomrule
  \end{tabular}
  }
  \end{minipage}
\begin{minipage}{0.39\linewidth}
  \caption{
      \textbf{Dataset comparison.}
      Our method is able to find more contact-rich, stable grasps compared to the GraspIt!~\cite{miller2004graspit} baseline and the training set of DDGC~\cite{lundell2021ddgc}. Specifically, we discover grasps with higher contact area (\textbf{CA}) and stability, as measured by epsilon metric ($\epsilon$), volume metric ($\textbf{Vol}$), and simulation displacement (\textbf{SD}).
      This results in somewhat higher interpenetration volume (\textbf{IV}), but maintains a similar ratio between intersection and contact area.
      All reported figures are top10 (ranked by epsilon metric).
  }
  \label{tab:2-exp1-datasets}
  \end{minipage}
\end{table*}

\begin{table*}[!t]
  \centering
  \begin{minipage}{0.6\linewidth}
  \resizebox{\linewidth}{!} 
  {
    \begin{tabular}{ll|S[table-format=4.5]S[table-format=4.5]S[table-format=4.5]S[table-format=4.5]S[table-format=4.5]}
  \toprule
  \rowcolor[HTML]{CBCEFB} 
  \textbf{Test Set} &\textbf{Train Set} &  \textbf{CA} $\uparrow$ & \textbf{IV} $\downarrow$ & \textbf{{\Large$\epsilon$} $\uparrow$} & \textbf{Vol $\uparrow$} & \textbf{SD $\downarrow$} \\ 
  \midrule
  \rowcolor{Lightapricot}
  Scale (Unit) & & \small{$\text{cm}^2$} & \small{$\text{cm}^3$} &  &  & \small{$\text{cm}$} \\ 
  \midrule
  EGAD Val. & \emph{DDGC}~\cite{lundell2021ddgc} & 3.17 & \textbf{12.37} & 0.11 & 0.35 & 1.09\\
  \rowcolor[HTML]{EFEFEF} 
  & \emph{GraspIt!}~\cite{miller2004graspit} & 3.30 & 13.62 & 0.15 & 0.71 & 1.48 \\
  & \emph{Ours} & \textbf{5.02} & 13.07 & \textbf{0.18} & \textbf{0.85} & \textbf{0.60} \\
  \midrule
  \rowcolor[HTML]{EFEFEF}
  KIT & \emph{DDGC} & 4.28 & \textbf{10.33} & 0.09 & 0.25 & 1.00 \\
  & \emph{GraspIt!} & 4.31 & 13.07 & 0.13 & 0.42 & 1.55 \\
  \rowcolor[HTML]{EFEFEF}
  & \emph{Ours} & \textbf{4.60} & 10.58 & \textbf{0.18} & \textbf{0.81} & \textbf{0.86} \\
  \midrule
  EGAD+KIT & \emph{DDGC} & 3.45 & 11.54 & 0.11 & 0.40 & 1.01\\
  \rowcolor[HTML]{EFEFEF}
  & \emph{GraspIt!} & 3.52 & 13.13 & 0.14 & 0.64 & 1.58\\
  & \emph{Ours} & \textbf{4.70} & \textbf{11.14} & \textbf{0.20} & \textbf{1.04} & \textbf{0.56} \\
  \bottomrule
  \end{tabular}
  }
  \end{minipage}
\begin{minipage}{0.39\linewidth}
  \caption{
  \textbf{Training vision-based grasping.}
  Retraining the DDGC~\cite{lundell2021ddgc} network,
  which predicts Barrett hand grasps from RGBD and instance segmentation inputs,
  with data generated by our method
  results in predicted grasps with 30\% more contact area (\textbf{CA}), 33\% higher epsilon metric ($\epsilon$) and 35\% lower simulated displacement (\textbf{SD}).
  }
  \label{tab:3-exp2-ddgc}
  \end{minipage}
\end{table*}

%
%

%% file: 5-experiments.tex
\section{Experiments}
\label{sec:experiments}
\paragraph{\textbf{Datasets}}
\emph{Grasp'D-Base} is our dataset of one million 
 \textit{base grasps} of canonically posed EGAD~\cite{morrison2020egad} and YCB~\cite{calli2017yale} objects with the Barrett, Allegro, and Shadow hands.
Our main dataset of multi-modal visual training examples, \emph{Grasp'D-1M}, is generated by transferring these base grasps to randomized scenes and rendering as described in Section~\ref{sec:method}.
In all experiments, we use the original DDGC~\cite{lundell2021ddgc} renders and scenes to match the original intended design parameters for a fair comparison.
The dataset labels used in Tables~\ref{tab:2-exp1-datasets} and~\ref{tab:3-exp2-ddgc} are explained below.
\emph{DDGC} refers to the Barrett hand dataset provided by~\cite{lundell2021ddgc}.
\emph{GraspIt!} refers to a set of baseline grasps for DDGC scenes generated with the GraspIt!~\cite{miller2004graspit} simulated annealing planner (for 70k steps for the Barrett and Allegro hands and 40k steps for the Shadow hand).
\emph{Ours} refers to a similar-sized set of grasps created by transferring grasps from \emph{Grasp'D-Base} to the same DDGC scenes.
\emph{MultiDex} refers to the simulation filtered grasp dataset provided by~\cite{li2022gendexgrasp} which we transfer to DDGC scenes in the same manner.
%
%

\paragraph{\textbf{Metrics}}
Grasps (dataset and predictions) are evaluated with geometric metrics:
(1) contact area (\textbf{CA}),
(2) intersection volume (\textbf{IV}),
(3) epsilon metric ($\epsilon$),
(4) grasp wrench space volume metric ($\textbf{Vol}$),
and
(5) simulation displacement ($\textbf{SD}$).
Notably, \textbf{CA}, $\epsilon$ and \textbf{Vol} depend on the distance threshold used for contact generation.
Previous works use varying thresholds (e.g., 9mm in~\cite{lundell2021ddgc} and 2mm in~\cite{grady2021contactopt}).
We plot these metrics for a wide range of thresholds (Figure~\ref{fig:3-grasp-metrics}) and show \modelName significantly outperforms baselines under all choices.

\subsection{\dataName Evaluation: Grasp Quality Metrics}
Figure~\ref{fig:3-grasp-metrics} shows that Shadow hand grasps generated by our method strictly dominate the GraspIt! and MultiDex baselines in terms
of epsilon quality, GWS volume, and contact surface area under all contact thresholds,
across two object sets (YCB and EGAD).
Results for the Barrett and Allegro hands (available on our website) follow a similar trend.
Figure~\ref{fig:qualitative} shows examples grasps from our method and GraspIt!.
Whereas GraspIt! mainly discovers grasps with fingertip contact, we find high-contact grasps that conform to object surface geometry.
Table~\ref{tab:2-exp1-datasets} shows additional statistics with a medium contact threshold of 5mm.

\subsection{RGBD Grasp Prediction with \dataName}
To confirm the generated data can in  practice improve vision-based grasp pipelines,
we retrain an existing network, DDGC~\cite{lundell2021ddgc}, on data from our method or the GraspIt! baseline.
For simplicity, we limit our evaluation to single-object scenes.
We follow the training procedure described in~\cite{lundell2021ddgc} and find training converges after 3000 epochs.
Table-\ref{tab:3-exp2-ddgc} reports our results, which show training on our dataset 
results in predicted grasps with 30\% more contact (\textbf{CA}), 33\% higher epsilon metric ($\epsilon$), and 35\% lower simulated displacement (\textbf{SD}).

\subsection{Computational Performance.}
We achieve a roughly 10x speedup compared to the GraspIt!~\cite{miller2004graspit} simulated annealing planner and a 20x speedup compared to a previous differentiable grasp synthesis method~\cite{graspd}.
The GraspIt! simulated annealing planner takes around 20s to generate a Barrett hand grasp.
~\cite{graspd} takes about 5 minutes to generate a single grasp online or 20 seconds amortized over parallel generations in a batch.
Our method can generate a contact-rich grasp of a YCB object in about 1s amortized, while online generation of a single grasp takes about 10s.
Grasp'D and \modelName timings are reported with a batch size of 32 on an NVIDIA A100.

%% file: 6-conclusion.tex
\section{Conclusion}
In this work, we have introduced a new method to synthesize multi-fingered grasps that leverages our differentiable grasping simulator to achieve improved stability and contact richness compared to commonly used baselines. Our experiments have demonstrated the benefits of our approach to an existing multi-finger robotic grasping pipeline that we trained on our generated grasps. With the release of a large-scale dataset of high-quality grasps synthesized through our method for various robotic hands, we aim to further advance the state of the art in multi-fingered robotic grasping.

\section{Acknowledgements}
The authors gratefully acknowledge the support of NSERC, Vector, CIFAR, UofT, Nvidia and Samsung.